%% file: main.tex
\documentclass[10pt,twocolumn,letterpaper]{article}

\usepackage[pagenumbers]{wacv} 
\usepackage{dblfloatfix}
\usepackage{multirow}

\definecolor{wacvblue}{rgb}{0.21,0.49,0.74}
\usepackage[pagebackref,breaklinks,colorlinks,allcolors=wacvblue]{hyperref}

\def\wacvPaperID{*****} 
\def\confName{WACV}
\def\confYear{2026}

\title{TOTNet: Occlusion-Aware Temporal Tracking for Robust Ball
Detection in Sports Videos}

\author{Hao Xu\\
Deakin University\\
Melbourne, Australia\\
{\tt\small august.xu@research.deakin.edu.au}
\and
Arbind Agrahari Baniya\\
Deakin University\\
Melbourne, Australia\\
{\tt\small a.agraharibaniya@deakin.edu.au}
\and
Sam Wells\\
Paralympics Australia\\
Melbourne, Australia\\
{\tt\small sam.wells@paralympic.org.au}
\and
Mohamed Reda Bouadjenek\\
Deakin University\\
Melbourne, Australia\\
{\tt\small reda.bouadjenek@deakin.edu.au}
\and
Richard Dazeley\\
Deakin University\\
Melbourne, Australia\\
{\tt\small richard.dazeley@deakin.edu.au}
\and
Sunil Aryal\\
Deakin University\\
Melbourne, Australia\\
{\tt\small sunil.aryal@deakin.edu.au}
}

\begin{document}

\twocolumn[{%
\renewcommand\twocolumn[1][]{#1}%
\maketitle
\includegraphics[width=1\linewidth]{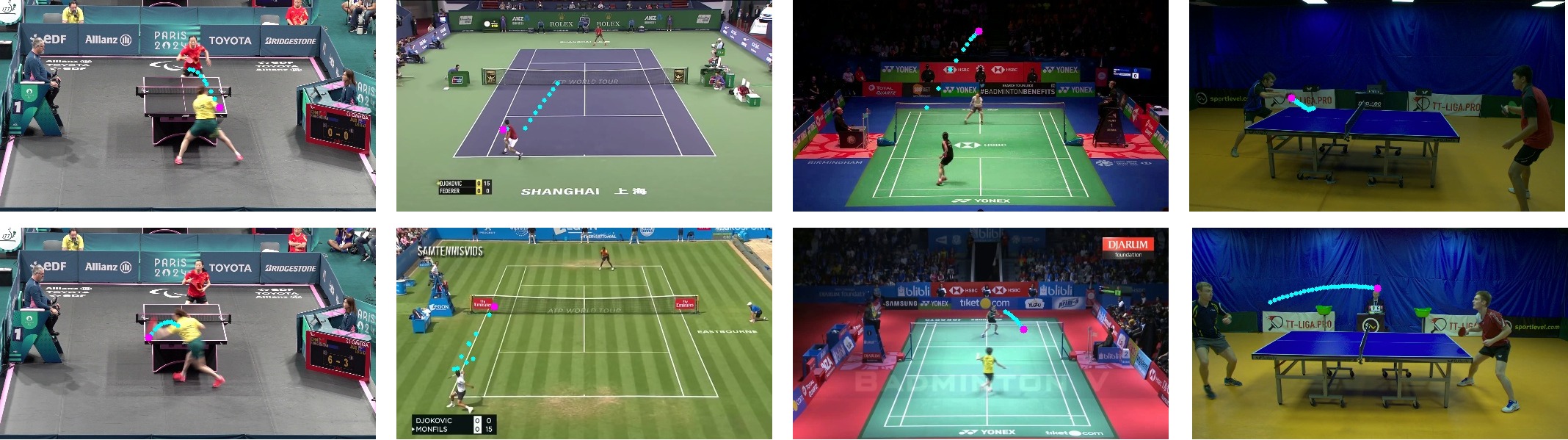}
\captionof{figure}{Tracking examples of TOTNet across four different sports videos, including scenarios with occlusion.}
\label{fig:teaser}
\vspace{1em} 
}]

\input{sec/0_abstract}    
\input{sec/1_intro}
\input{sec/2_lit_review}
\input{sec/3_methods}
\input{sec/4_experiments}
\input{sec/5_conclusion}
{
    \small
    \bibliographystyle{ieeenat_fullname}
    \bibliography{main}
}

\end{document}

%% file: sec/0_abstract.tex
\begin{abstract}
   Robust ball tracking under occlusion remains a key challenge in sports video analysis, affecting tasks like event detection and officiating. We present TOTNet, a Temporal Occlusion Tracking Network that leverages 3D convolutions, visibility-weighted loss, and occlusion augmentation to improve performance under partial and full occlusions. Developed in collaboration with Paralympics Australia, TOTNet is designed for real-world sports analytics. We introduce TTA, a new occlusion-rich table tennis dataset collected from professional-level Paralympic matches, comprising 9,159 samples with 1,996 occlusion cases. Evaluated on four datasets across tennis, badminton, and table tennis, TOTNet significantly outperforms prior state-of-the-art methods, reducing RMSE from 37.30 to 7.19 and improving accuracy on fully occluded frames from 0.63 to 0.80. These results demonstrate TOTNet’s effectiveness for offline sports analytics in fast-paced scenarios. Code and data access:\href{https://github.com/AugustRushG/TOTNet}{AugustRushG/TOTNet}.
\end{abstract}

%% file: sec/1_intro.tex
\section{Introduction}
\label{sec:intro}

Deep learning-based computer vision techniques have revolutionised sports analytics, enabling applications such as performance analysis, automated officiating, and player tracking \cite{naik2022comprehensive}. One of the fundamental tasks in this domain is ball tracking, which involves continuously locating the position of a ball in a sequence of video frames \cite{kamble2019ball}. The task is crucial for understanding game dynamics, as it enables downstream applications such as ball possession analysis, trajectory prediction, and event detection (e.g., shot position). Accurate ball tracking requires handling challenges such as the ball's small size, rapid movement, and frequent occlusions.

However, tracking fast-moving balls under occlusion remains a significant challenge \cite{halbinger2015video,naik2023yolov3}, in a single-view set-up. Occlusions caused by players, equipment, or environmental factors disrupt crucial tasks, such as ball possession analysis, key pass tracking, and scoring event detection. These challenges affect coaching decisions, player performance metrics, and game outcomes, making reliable ball tracking a fundamental need.

Traditional object trackers, inspired by state-of-the-art (SOTA) detectors such as You Only Look Once (YOLO) \cite{redmon2016you}, Faster R-CNN \cite{ren2016faster}, and Single Shot Detector (SSD) \cite{liu2016ssd}, face significant challenges under occlusion conditions \cite{saleh2021occlusion}. These methods rely heavily on spatial data from individual frames, making them inadequate for predicting ball locations when the ball becomes temporarily invisible due to occlusion.
Existing ball tracking methods utilizing Convolutional Neural Networks (CNNs) \cite{liu2022monotrack,huang2019tracknet,sun2020tracknetv2,tarashima2023widely,voeikov2020ttnet} incorporate temporal information by stacking multiple frames together as input. However, this approach is inherently limited in capturing complex temporal dependencies and motion dynamics, which are critical for accurately tracking fast-moving and occluded objects. Stacking frames treats temporal information as static features rather than dynamic sequences, losing critical inter-frame relationships. Additionally, this method is unable to explicitly model motion patterns or non-linear trajectories, which are common in sports.
Other methods \cite{naik2023yolov3,hu2024basketball,li2023table} address the occluded ball tracking problem by employing Kalman Filters (KF) or similar techniques, which estimate the ball's optimal state in the current frame based on its previous states. However, KF is inherently designed for linear systems, making it unsuitable for tracking non-linear and dynamically moving objects, such as balls with erratic trajectories caused by spin, sudden deflections, or rapid changes in velocity. Moreover, the integration of Kalman filters adds complexity to the system, functioning as a post-processing step \cite{kamble2019deep}.

\begin{figure}
    \centering
    \includegraphics[width=1\linewidth]{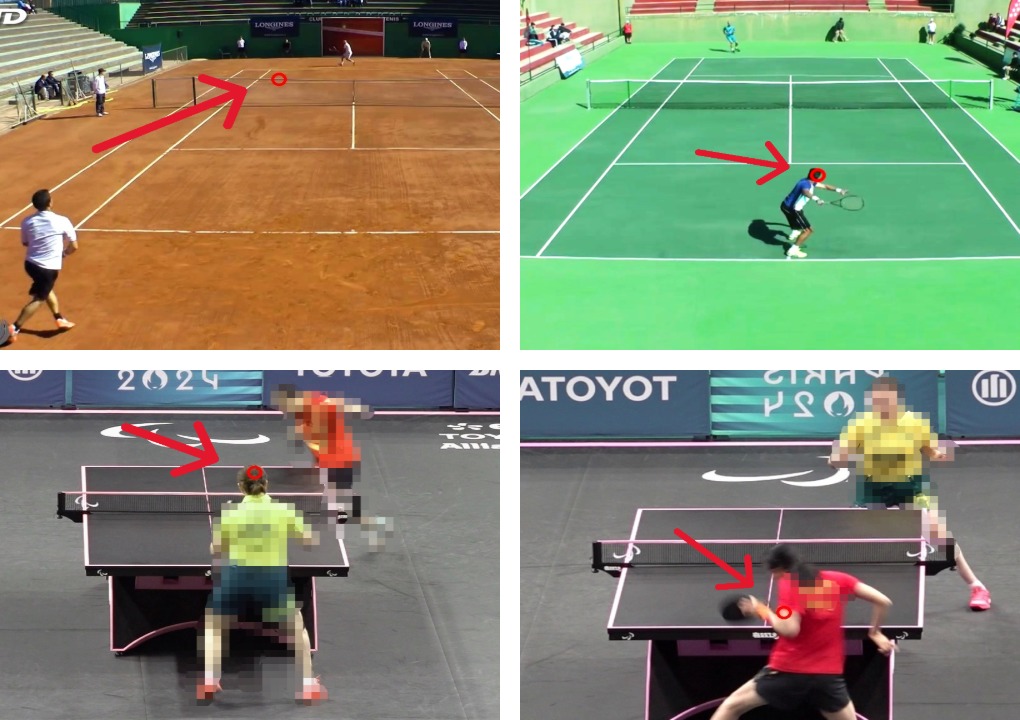}
    \caption{Occlusion Examples where the ball is drawn with a red circle.}
    \label{fig:occlusion_examples}
\end{figure}

Another major limitation is the lack of diversity in publicly available sports ball tracking datasets. These datasets often fail to represent the wide variety of occlusion scenarios encountered in real-world conditions, leading to models that struggle to generalise effectively. Additionally, low frame rates in many recordings exacerbate challenges such as motion blur, further degrading detection accuracy, as illustrated in Figure~\ref{fig:occlusion_examples}.

In fast-paced sports such as tennis, badminton, and table tennis, occlusions occur frequently, underscoring the need for robust tracking methods that can effectively leverage both temporal and contextual information. Beyond sports analytics, addressing the challenges of tracking occluded objects has broader implications for other domains, such as autonomous driving, where vehicles or pedestrians may become temporarily hidden. Developing solutions that improve ball tracking under occlusion will not only advance sports analytics but also enhance applications in safety-critical domains.

To address these challenges, we propose TOTNet (Temporal Occlusion Tracking Network), a novel framework that integrates spatial and temporal information using specialised learning modules, occlusion-aware data augmentation. Unlike prior work, TOTNet is explicitly designed for \textit{offline sports analytics} applications such as post-match analysis, referee support, and performance breakdowns—where inference speed is less critical than accurate and robust tracking under occlusion. This enables TOTNet to focus on achieving higher accuracy and generalisation across diverse sports and visibility conditions.
TOTNet achieves state-of-the-art performance, substantially improving occluded ball tracking across diverse sports datasets, as illustrated in Figure~\ref{fig:teaser}. Key contributions of this work include:
\begin{itemize}
    \item \textbf{TTA: An occlusion-rich table tennis tracking dataset}: We introduce the TTA dataset, a manually annotated table tennis dataset featuring 9,159 samples, including 1,996 occlusion cases. Designed to reflect real-world gameplay conditions, it provides a valuable benchmark for evaluating object tracking performance under challenging occlusion scenarios.
    \item \textbf{Novel occlusion augmentation technique}: A data augmentation strategy that simulates occlusions by masking the ball area in target frames, forcing the model to rely on temporal and spatial context rather than solely spatial features.
    \item \textbf{Integration of temporal and spatial information}: A novel architecture that retains the temporal dimension and incorporates a specially designed temporal information learning module, enabling effective utilisation of motion dynamics for occluded object tracking. 
\end{itemize}
Our model TOTNet achieves SOTA performance, significantly improving occluded ball tracking on the TTA dataset by reducing RMSE from 37.30 to 7.19 and increasing accuracy from 0.63 to 0.80. Additionally, in the tennis dataset \cite{huang2019tracknet}, it demonstrates remarkable improvements in tracking hard-to-detect objects, reducing RMSE from the previous SOTA 105.73 to 63.41.

%% file: sec/2_lit_review.tex
\section{Related Work}
\label{sec:lit_rev}
\subsection{Single Object Tracking in Sports Videos}
The development of deep learning-based image detectors such as YOLO \cite{redmon2016you}, SSD \cite{liu2016ssd}, and R-CNN \cite{girshick2014richfeaturehierarchiesaccurate} has significantly advanced ball tracking in sports videos. These methods follow the tracking-by-detection (TBD) paradigm, where detections are obtained from individual frames and subsequently linked to form trajectories \cite{naik2023yolov3,buric2018ball,teimouri2019real,reno2018convolutional,komorowski2019deepball}. However, TBD methods process frames independently, which limits their ability to leverage temporal information and results in temporally inconsistent tracking, especially during partial or full occlusions.

To overcome these limitations, recent works have explored the integration of temporal information. Methods like TrackNet \cite{huang2019tracknet}, TrackNetV2 \cite{sun2020tracknetv2}, and MonoTrack \cite{liu2022monotrack} incorporate multiple consecutive frames as inputs to CNNs, capturing short-term motion patterns. Other approaches use advanced temporal modelling techniques, such as optical flow \cite{dosovitskiy2015flownet}, Recurrent Neural Networks (RNNs), convolutional LSTMs \cite{patraucean2015spatio}, and temporal convolutions \cite{lea2017temporal}, to better model object motion over time \cite{kukleva2019utilizing,li2023table}. Additionally, transformers \cite{vaswani2017attention} have introduced spatiotemporal attention mechanisms, enabling models to learn correlations within and across frames and predict object movements more effectively \cite{yu2024towards,chao2024tracking}.
Despite these advances, current approaches still struggle with occlusion handling, particularly in sports where the ball frequently disappears due to rapid motions and interactions with players. This highlights the need for methods that can effectively leverage both temporal and contextual information to improve tracking consistency and its robustness to occlusion challenges.

\subsection{Occlusion Object Tracking}
Occluded object tracking remains a significant challenge in video-based object detection, despite advancements in the field \cite{saleh2021occlusion}. The difficulty lies in collecting and labelling datasets with sufficient occlusion diversity, as creating comprehensive real-world datasets for all occlusion scenarios is nearly impossible. As a result, many studies are based on synthetic datasets or automatically generated occluded samples \cite{saleh2021occlusion}.
To address this, Generative Adversarial Networks (GANs) \cite{goodfellow2014generative} have been employed to generate occluded data. For instance, \cite{wang2017fast} augmented the COCO dataset with occluded objects using GANs, improving model robustness through enhanced training data. Similarly, \cite{li2016amodal} created synthetic occlusions by overlaying object masks from one image onto another, producing amodal data to improve occlusion handling.

Compositional models also show promise. These models detect partially occluded objects by leveraging a generative, modular approach. For example, \cite{kortylewski2020compositional} used a differentiable generative compositional layer instead of the fully connected layer in a CNN, enabling robust classification of occluded objects and accurate localisation of occluders.
In another approach, \cite{cui2021remote} framed object tracking as a Markov decision process within a deep reinforcement learning framework. Their AD-OHNet tracker utilised temporal and spatial contexts from action-state histories prior to occlusion, enabling accurate tracking even during complete occlusion.
For multiperson tracking, \cite{zhou2018deep} proposed a deep alignment network that combines an appearance model with a Kalman filter-based motion model. This hybrid approach effectively handles occlusions and improves motion reasoning during tracking.

Occlusion is a common challenge in sports scenarios, significantly impacting ball and player detection. For instance, \cite{halbinger2015video} introduced a two-stage approach for detecting soccer balls under occlusion. The first stage detects the ball when fully visible, while the second stage identifies occluded parts as "bumps" on player silhouettes using a Hough transform for circular shape detection, followed by Freeman Chain Code analysis to confirm the ball's dimensions.
In \cite{kamble2019deep}, the authors tackled ball occlusion by tracking the player occluding the ball, maintaining continuity in the tracking process. However, this post-processing approach is less effective in sports with frequent and complex occlusions that demand precise, real-time localisation.
To address occlusions, \cite{naik2023yolov3} integrated YOLOv3 with a Kalman filter to predict the ball's position during occlusion. This hybrid approach demonstrated improved robustness in challenging scenarios by combining detection and prediction.
In racket sports, leveraging temporal information is critical due to frequent occlusions of fast-moving objects. For example, \cite{huang2019tracknet} demonstrated that by stacking multiple consecutive frames as input enables CNN models to capture trajectory patterns, aiding in the detection and tracking of occluded objects. However, the mechanisms by which trajectory patterns are learned remain unexplored and the occlusion samples in the dataset is too small to show its effectiveness. 
Building on this, \cite{sun2020tracknetv2} improved prediction accuracy by adopting a U-Net structure and a multiple-input, multiple-output (MIMO) framework. While this approach slightly reduced processing speed, it enhanced overall performance.
Further improvements were made by \cite{liu2022monotrack}, who incorporated residual connections within U-Net blocks to enhance tracking performance in badminton videos. 
Contrasting with encoder-decoder architectures, \cite{tarashima2023widely} argued that such designs often lack sufficient feature diversity for effective tracking. They employed high-resolution feature extraction methods from HRNet \cite{wang2020deep,yu2021lite}, combined with position-aware model training and temporal consistency, achieving SOTA results across multiple sports datasets.

Most existing approaches leverage temporal information by stacking multiple frames along the channel dimension for 2D convolutions. However, this method limits the ability to fully capture the rich temporal dynamics inherent in video data. Stacking frames treats temporal information as static features rather than dynamic sequences, failing to explicitly model the evolution of object motion over time. As a result, these approaches struggle to track objects accurately during occlusion, where the model must rely on contextual information from preceding and succeeding frames to predict the occluded object's position.

%% file: sec/3_methods.tex
\section{Methodology}
\label{sec:methods}
\subsection{Datasets}
\label{Dataset} 
To evaluate performance across varied racket sports, we use four datasets: three existing ones for table tennis, tennis, and badminton, and a newly introduced TTA dataset designed to emphasize occlusion scenarios.
The TT dataset from \cite{voeikov2020ttnet} includes five training and seven testing videos, yielding 36,224 training, 3,232 validation, and 3,720 test samples. Captured at 120 fps ($1080\times1920$) from a side view, it features minimal occlusion and lacks visibility labels.
The tennis dataset, accessed via \cite{tarashima2023widely} from \cite{huang2019tracknet}, contains 10 clips (30 fps, $720\times1080$) with ball coordinates and visibility labels: 0 (out-of-frame), 1 (visible), 2 (partially visible), 3 (fully occluded). We created balanced splits detailed in Table~\ref{tab:combined_visibility_dataset}.
The badminton dataset \cite{sun2020tracknetv2} has 26 training and 3 testing matches (30 fps, $720\times1280$) with binary visibility labels (0 = not visible, 1 = visible). It contains a higher proportion of invisible frames than other datasets.

\begin{table}[htbp]
    \centering
    \renewcommand{\arraystretch}{1.0} 
    \resizebox{1\linewidth}{!}{
    \begin{tabular}{|l|l||c|c|c|}
        \hline
        \textbf{Dataset} & \textbf{Visibility Level} & \textbf{Train} & \textbf{Val} & \textbf{Test} \\
        \hline\hline
        \multirow{4}{*}{Tennis}
            & Out of Frame        & 587 & 107 & 135 \\
            & Fully Visible       & 11,641 & 2,940 & 3,054 \\
            & Partially Occluded  &   861  &  193  &  338  \\
            & Fully Occluded      &    55  &   22  &    5  \\
        \hline
        \multirow{3}{*}{\textbf{TTA (ours)}}
            & Fully Visible        & 5,141  & 1,285 &  668 \\
            & Partially Occluded  &   834  &  215  &   72 \\
            & Fully Occluded      &   650  &  158  &   67 \\
        \hline
        \multirow{2}{*}{Badminton}
            & Out of Frame         & 8,052  & 2,022 & 2,188 \\
            & Fully Visible        & 54,794 &13,690 &10,468 \\
        \hline
    \end{tabular}
    }
    \caption{Visibility-level distribution across Tennis, TTA, and Badminton datasets.}
     \label{tab:combined_visibility_dataset}
    \renewcommand{\arraystretch}{1.0} 
\end{table}

\begin{figure}
    \centering
    \includegraphics[width=0.9\linewidth]{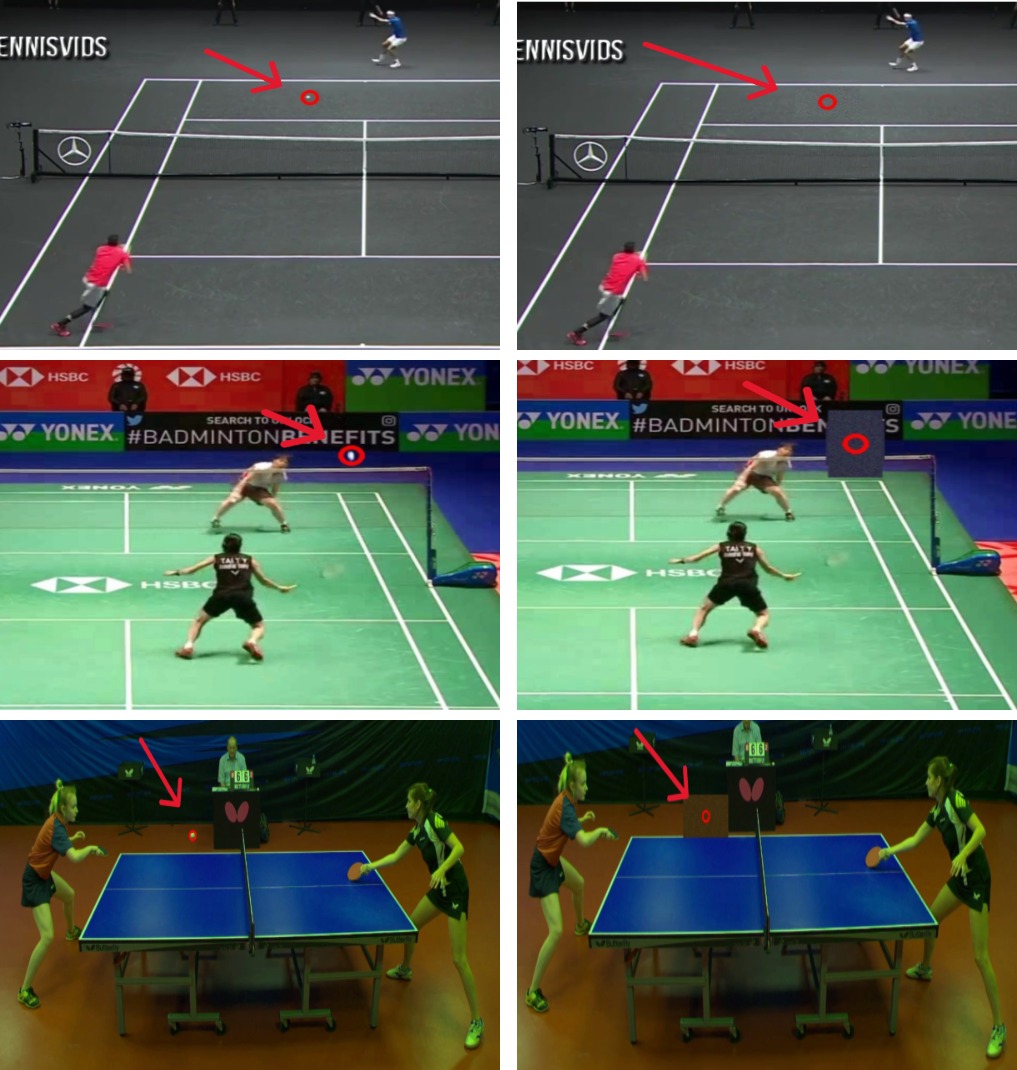}
    \caption{Augmentation Examples for three different datasets where the ball is circled with red}
    \label{fig:augmentationl}
\end{figure}

\begin{figure*}[t!]
    \centering
    \includegraphics[width=0.95\textwidth]{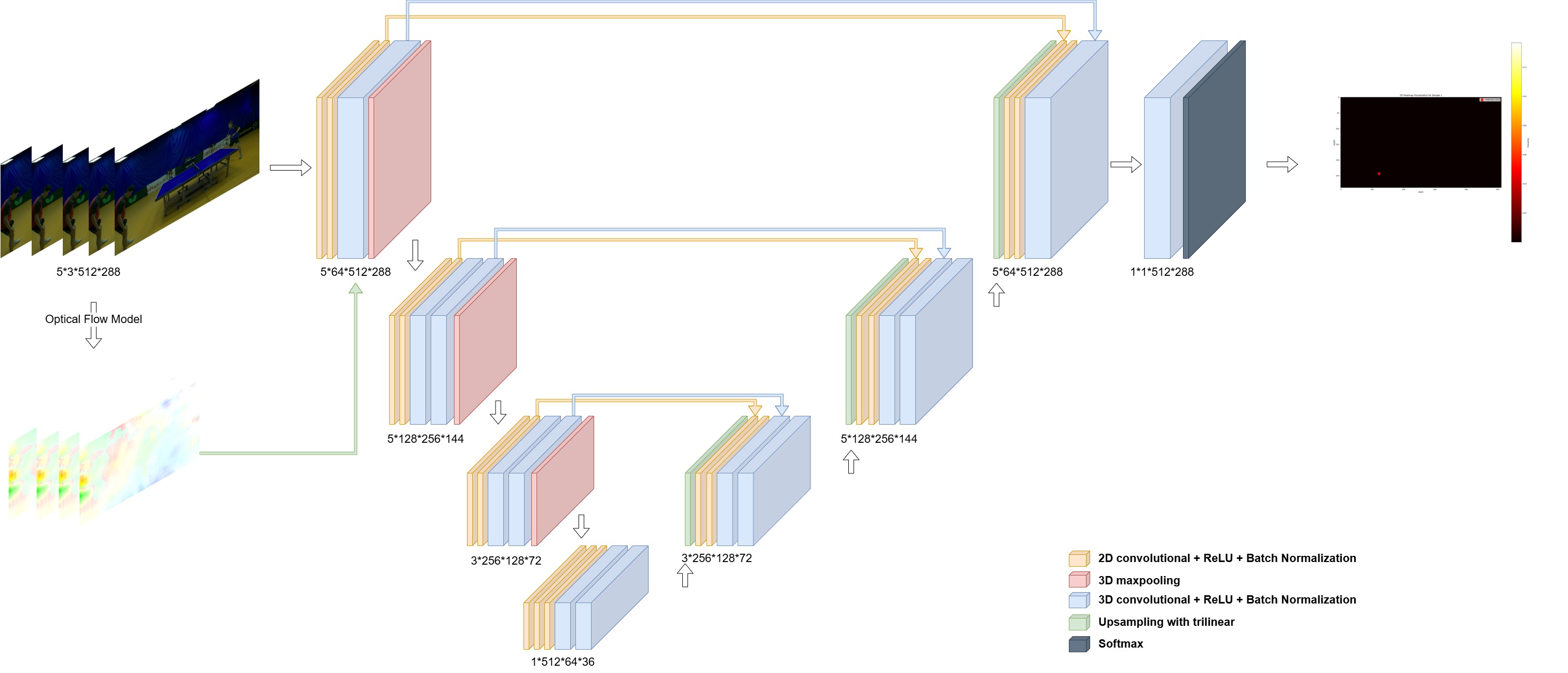}
    \caption{Overview of the architecture design.}
    \label{fig:architecture}
\end{figure*}

The \textbf{TTA dataset}, manually annotated for this study, includes 9,159 samples (25 fps, $1080\times1920$) from four professional-level Paralympic table tennis matches. Annotated following standard visibility labels \cite{huang2019tracknet,sun2020tracknetv2}, it includes 1,996 occlusion samples—more than any existing dataset—due to camera angles, ball size, and dynamic play. Labels were reviewed with national team analysts to ensure quality. Designed for real-world occlusion benchmarking, sample frames are shown in Figure~\ref{fig:occlusion_examples}. The dataset will be shared upon academic request.

\textbf{Note:} Out-of-frame samples in the tennis dataset are excluded from training due to their rarity and negligible impact; they are filtered during inference via confidence thresholds.

\subsection{Occlusion Augmentation}
We propose a novel augmentation technique to enhance performance in occlusion scenarios. For a sequence of frames, we simulate occlusion by masking the ball area in the target frame with a randomly sized shape filled with the surrounding mean pixel values. Examples are shown in Figure \ref{fig:augmentationl}. This augmentation forces the model to rely on temporal information from adjacent frames and the spatial context surrounding the occluded region rather than solely depending on the current frame's spatial features.

To prevent the model from adapting too strongly to this augmentation, we introduce additional noise by randomly selecting areas in the other frames and filling them with mean pixel values. This ensures that the model learns to generalise better and robustly utilise both temporal and spatial features across all frames.

\subsection{BCE Loss with Visibility-Based Weighting}

We introduce a visibility-aware weighted binary cross-entropy (BCE) loss to address imbalances and ambiguities inherent in occlusion-heavy tracking scenarios. Prior approaches~\cite{huang2019tracknet,sun2020tracknetv2,tarashima2023widely,voeikov2020ttnet} apply Gaussian target maps uniformly, but we found this does not consistently enhance convergence. Moreover, assigning a fixed coordinate (e.g., (0,0)) for out-of-frame instances can introduce a biased loss, leading the model to overpredict such cases, especially when they dominate the dataset.

Our formulation addresses this by:
\begin{itemize}
    \item Representing out-of-frame cases as explicit "no-target" signals without Gaussian maps.
    \item Using normalized Gaussian targets for fully occluded frames to encode positional uncertainty.
    \item Assigning visibility-dependent weights to ensure balanced learning across all visibility levels.
\end{itemize}

\paragraph{Target Definition.}
Let $P_x \in \mathbb{R}^{W}$ and $P_y \in \mathbb{R}^{H}$ be the predicted 1D heatmaps along the width and height axes. Given ground truth coordinates $T_x, T_y \in \mathbb{R}$, the targets are defined as:

- \textbf{Visible or Partially Occluded (levels 1–2):} One-hot encoding:
\[
T_{x,\text{map}}[j] = \begin{cases} 
1, & j = T_x \\
0, & \text{otherwise}
\end{cases}, \quad
T_{y,\text{map}}[k] = \begin{cases} 
1, & k = T_y \\
0, & \text{otherwise}
\end{cases}
\]

\begin{align}
T_{x,\text{map}}[j] &= \frac{1}{Z_x} \exp\left(-\frac{(j - T_x)^2}{2\sigma^2}\right), \\
T_{y,\text{map}}[k] &= \frac{1}{Z_y} \exp\left(-\frac{(k - T_y)^2}{2\sigma^2}\right)
\end{align}
where $Z_x$ and $Z_y$ normalize the distributions to sum to 1.

\paragraph{Visibility-Based Weighting.}
Each visibility level $v \in \{0, 1, 2, 3\}$ is associated with a weight $w_v$, defined via a vector $\mathbf{w} = [w_0, w_1, w_2, w_3]$.

\paragraph{Final Loss.}
The loss for a given instance is:
\[
L = w_v \cdot \left( \text{BCE}(P_x, T_{x,\text{map}}) + \text{BCE}(P_y, T_{y,\text{map}}) \right)
\]

\subsection{Architecture}
Our model extends prior work \cite{huang2019tracknet,sun2020tracknetv2} by preserving the temporal dimension, enabling richer motion-aware feature learning. Unlike earlier methods that stack frames along the channel axis, we retain temporal structure throughout the network. Built on a U-Net backbone \cite{ronneberger2015u}, the model uses encoder-decoder blocks with skip connections for spatial detail, and integrates optical flow from a pre-trained RAFT model \cite{teed2020raft} in the initial encoder to enhance motion sensitivity. The architecture is shown in Figure~\ref{fig:architecture}.

\subsubsection{Encoder}
\begin{figure}
    \centering
    \includegraphics[width=\linewidth]{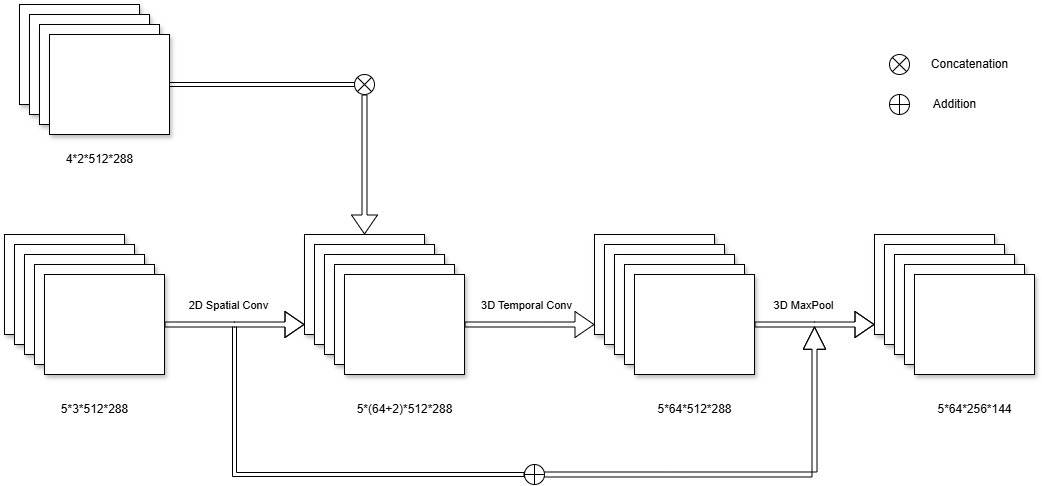}
    \caption{The specification of the first encoder block}
    \label{fig:encoder_block}
\end{figure}

Each encoder block integrates spatial and temporal convolutions, where the spatial convolution is implemented as a 2D convolution, and the temporal convolution is implemented as a 3D convolution. Initially, spatial convolutions are applied to each frame independently to extract spatial features. These features are then passed through temporal convolutions, keeping the temporal dimension intact, to capture dependencies across frames and effectively combine spatial and temporal information for better object localisation. A 3D max pooling operation is then applied to reduce both the spatial size and temporal frame size while preserving the most valuable information. Additionally, a residual connection is included within each encoder block, where the output from the spatial convolution is added to the output of the temporal convolution to ensure that spatial information is not lost. A detailed flow of the encoder block is specified in Figure \ref{fig:encoder_block}.
As the model deepens on the encoder side, the number of channels increases while the spatial resolution and temporal sequence decrease. The kernel size for spatial convolutions becomes smaller to focus on finer details in smaller spatial regions, whereas the kernel size for temporal convolutions decreases as the temporal dimension gets reduced. 

\subsubsection{Bottleneck}

The bottleneck block processes the highly abstracted information from the encoder. By the time the input reaches the bottleneck block, the temporal dimension has been reduced to 1, and the spatial dimensions have been significantly downsized. This block contains more spatial layers than the other blocks, focusing on extracting the most critical features.
Since the temporal dimension is reduced to 1, the kernel size for temporal convolution is set to (1,1,1), effectively performing point-wise convolution. This operation facilitates mixing information across all channels, enabling the block to generate rich feature representations that combine temporal and spatial information effectively.

\begin{table*}
    \centering
    \renewcommand{\arraystretch}{1.0}
    \resizebox{\textwidth}{!}{
    \begin{tabular}{|l|l||c|c|c||c|c|c||c|c||c||c|}
    \hline
    \multirow{2}{*}{Model} & \multirow{2}{*}{Metric} 
    & \multicolumn{3}{c||}{Tennis} 
    & \multicolumn{3}{c||}{TTA} 
    & \multicolumn{2}{c||}{Badminton} 
    & TT (Overall) 
    & Parameters M / FPS \\
    \cline{3-10}
    & & Visible & Partial Occ. & Full Occ. 
      & Visible & Partial Occ. & Full Occ. 
      & Visible & Not Visible 
      & & \\
    \hline
    \multirow{2}{*}{TTNet \cite{voeikov2020ttnet}} 
    & RMSE      & 25.40 & 72.70 & 48.77 & 16.31 & 18.38 & 17.23 & 40.76 & 43.31 & 4.02 & 7.62 / 40.75\\
    & Accuracy  & 0.23  & 0.19  & 0.17  & 0.20  & 0.10  & 0.27  & 0.23  & 0.74  & 0.85 & \\
    \hline
    \multirow{2}{*}{TrackNetV2 \cite{sun2020tracknetv2}} 
    & RMSE      & 24.48 & 109.93 & 232.70 & 3.20 & 13.65 & 100.89 & 34.32 & 32.95 & 2.43 & 11.34 / 40.74\\
    & Accuracy  & 0.84  & 0.49   & 0.00   & 0.96 & 0.88  & 0.51   & 0.86  & 0.88  & 0.91 & \\
    \hline
    \multirow{2}{*}{monoTrack \cite{liu2022monotrack}} 
    & RMSE      & 43.28 & 138.39 & 227.71 & 3.30 & 7.42  & 39.55  & 40.13 & \textbf{27.67} & 2.23 & 2.84 / 44.67\\
    & Accuracy  & 0.78  & 0.42   & 0.00   & 0.97 & 0.85  & 0.67   & 0.84  & \textbf{0.90} & 0.95 & \\
    \hline
    \multirow{2}{*}{WASB \cite{tarashima2023widely}} 
    & RMSE      & 16.58 & 105.73 & 264.45 & 2.14 & 5.26  & 37.30  & 27.17 & 56.50 & 3.11 & \textbf{1.48} / 33.44\\
    & Accuracy  & 0.92  & 0.52   & 0.17   & 0.97 & 0.88  & 0.63   & 0.88  & 0.79  & 0.84 & \\
    \hline
    \multirow{2}{*}{TOTNet} 
    & RMSE      & \textbf{6.07} & \textbf{63.41} & 27.98 & 2.19 & 3.79 & 12.31 & \textbf{23.43} & 44.55 & 1.67 & 8.65 / 28.08\\
    & Accuracy  & \textbf{0.95} & \textbf{0.61}  & \textbf{0.67} & 0.97 & 0.89 & 0.74 & \textbf{0.89} & 0.84 & 0.97 & \\
    \hline
    \multirow{2}{*}{TOTNet (OF)} 
    & RMSE      & 9.59 & 67.37 & \textbf{15.31} & \textbf{1.84} & \textbf{3.45} & \textbf{7.19} & 28.25 & 70.22 & \textbf{1.38} & 8.66 / 12.19\\
    & Accuracy  & 0.92 & 0.53  & 0.33 & \textbf{0.98} & \textbf{0.92} & \textbf{0.80} & 0.85 & 0.75 & \textbf{0.98} & \\
    \hline
    \end{tabular}
    }
    \caption{Performance comparison across Tennis, TTA, and Badminton datasets. RMSE and Accuracy are reported per visibility level. The rightmost column reports model efficiency in terms of parameter count (M), average inference FPS. TOTNet refers to the model without optical flow; TOTNet (OF) includes optical flow.}
    \label{tab:combined_results}
\end{table*}

\subsubsection{Decoder}
\begin{figure}
    \centering
    \includegraphics[width=\linewidth]{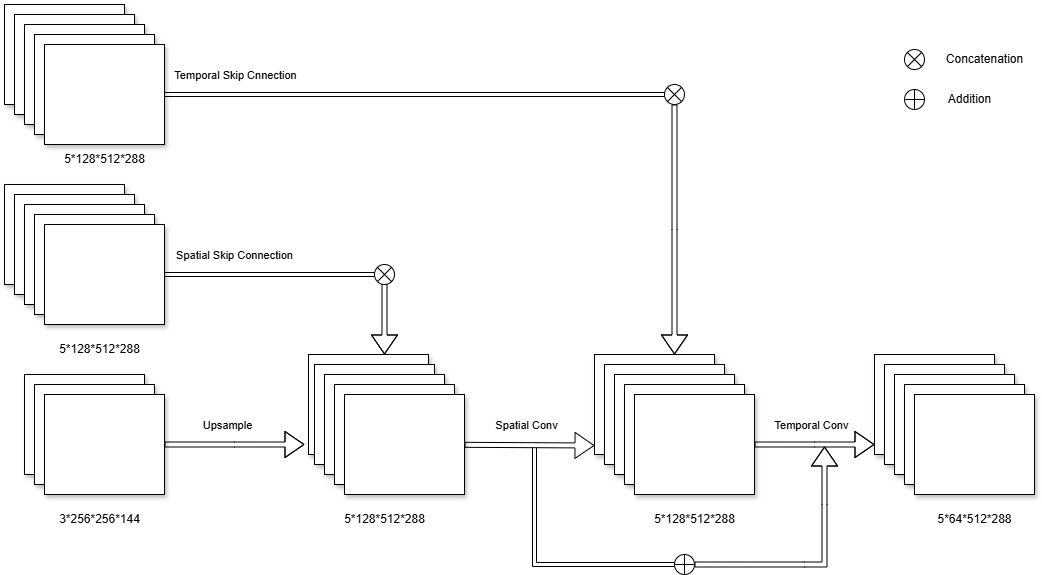}
    \caption{The specification of the last decoder block}
    \label{fig:decoder_block}
\end{figure}

After passing through the bottleneck block, the decoder blocks begin the 3D upsampling process to restore both the temporal and spatial dimensions. Features are upsampled using the trilinear method, as the model focuses on generating a heatmap for the object's likely position rather than precise segmentation. Each decoder block mirrors the layers of the encoder block, and skip connections are utilised for both spatial and temporal convolutions. These skip connections concatenate the encoder features along the channel dimension, allowing the decoder to leverage information from earlier layers for enhanced feature reconstruction both spatially and temporally. A detailed decoder structure is in Figure \ref{fig:decoder_block}.
In the final stage, a temporal convolutional block is employed to reduce the channel dimension to one, retaining the most critical information. A softmax activation is applied to the resulting heatmap to ensure it represents a normalised probability distribution.

%% file: sec/4_experiments.tex
\section{Experiments}
\label{sec:experiments}
For all datasets, video frames and corresponding ball coordinates are resized to $288\times512$ pixels. The number of input frames is a configurable hyperparameter, with five frames identified through experimentation as the optimal balance between temporal context and computational efficiency. In addition to occlusion augmentation, standard data augmentations such as color jittering, random cropping and resizing, and horizontal/vertical flipping are applied. The model is trained using the AdamW optimizer \cite{loshchilov2017decoupled}, with detailed hyperparameters and training settings provided in the GitHub repository. We evaluate two variants: the baseline TOTNet and TOTNet(OF), which incorporates optical flow features extracted via the pretrained RAFT model \cite{teed2020raft}.

\subsection{Evaluation}
We evaluate performance using RMSE and accuracy. RMSE captures the average squared distance between predicted and ground truth coordinates, while accuracy reflects the percentage of predictions within a defined threshold:

\begin{equation}
    \text{dist} = \sqrt{(x_{\text{pred}} - x_{\text{label}})^2 + (y_{\text{pred}} - y_{\text{label}})^2},
\end{equation}

Predictions within this threshold are considered correct. For fully visible and partially occluded objects, the threshold is set to 5 pixels, approximating the ball’s average size. For fully occluded objects, a relaxed threshold of 10 is used to accommodate annotation uncertainty.

\begin{table}[t]
    \centering
    \caption{Ablation study on the TTA dataset across different visibility levels. RMSE and Accuracy are reported separately for each method. WBCE = Weighted Binary Cross Entropy, Aug. = Occlusion Augmentation, OF. = Optical Flow.}
    \label{tab:ablation_study}
    \vspace{0.5em}
    \small
    \renewcommand{\arraystretch}{1.5}
    \resizebox{1\linewidth}{!}{
        \begin{tabular}{|l|c|c|c||c|c|c|}
            \hline
            \textbf{Method} & \textbf{WBCE} & \textbf{Aug.} & \textbf{OF.} & \textbf{Visible} & \textbf{Part. Occ.} & \textbf{Full Occ.} \\
            \hline \hline
            \multicolumn{7}{|c|}{\textbf{RMSE (↓ is better)}} \\
            \hline
            TOTNet (Baseline)             & --         & --         & --         & 2.57 & 6.38  & 29.57 \\
            TOTNet (WBCE)                & \checkmark & --         & --         & 2.36 & 5.72  & 24.43 \\
            TOTNet (Aug.)                & --         & \checkmark & --         & 2.67 & 12.20 & 54.26 \\
            TOTNet (WBCE + Aug.)         & \checkmark & \checkmark & --         & 2.19 & 3.79  & 12.31 \\
            TOTNet (WBCE + Aug. + OF.)   & \checkmark & \checkmark & \checkmark & \textbf{1.84} & \textbf{3.45} & \textbf{7.19} \\
            \hline
            \multicolumn{7}{|c|}{\textbf{Accuracy (↓ is better)}} \\
            \hline
            TOTNet (Baseline)             & --         & --         & --         & 0.97 & 0.85  & 0.54 \\
            TOTNet (WBCE)                & \checkmark & --         & --         & 0.97 & 0.86  & 0.61 \\
            TOTNet (Aug.)                & --         & \checkmark & --         & 0.97 & 0.83  & 0.56 \\
            TOTNet (WBCE + Aug.)         & \checkmark & \checkmark & --         & 0.97 & 0.89  & 0.74 \\
            TOTNet (WBCE + Aug. + OF.)   & \checkmark & \checkmark & \checkmark & \textbf{0.98} & \textbf{0.92} & \textbf{0.80} \\
            \hline
        \end{tabular}
    }
\end{table}

\subsection{Other Models}
In this work, we re-implemented models from \cite{sun2020tracknetv2}, \cite{liu2022monotrack,tarashima2023widely}, and \cite{voeikov2020ttnet} due to the lack of publicly available implementations. Details of re-implementation and adaptations, including resolution considerations for all models listed, are provided in the GitHub repository. Additionally \cite{tarashima2023widely} represents the most recent and advanced model for ball tracking tasks based on their superior performance comparing to all other models, serving as a strong benchmark for comparison.

\subsection{Results}
The results for all four datasets are presented in Table \ref{tab:combined_results}. Our proposed model, TOTNet, consistently outperforms current SOTA models across all datasets, particularly in handling occluded objects.
In the tennis dataset, for the partially occluded objects, TOTNet significantly reduced the RMSE from 105.73 to 63.41. More impressively, for fully occluded objects, it achieved an RMSE of 15.31 and improved the accuracy to 0.67, showcasing its ability to accurately predict the trajectory of the object despite complete occlusion. In the TTA dataset, TOTNet further demonstrated its effectiveness under occlusion settings, outperforming all other models across all visibility levels. Notably, it achieved an accuracy of 0.80 and an RMSE of 7.19 for fully occluded objects, underscoring its robustness in challenging scenarios. These results highlight the model's ability to leverage temporal and spatial information from previous frames to make accurate predictions.
In the TT and badminton datasets, TOTNet also surpassed all SOTA models across all metrics, showcasing its ability to handle both clear and occluded targets effectively. In the TT dataset, TOTNet reduced the RMSE from 4.02 to 1.38 and improved the accuracy to 0.98, significantly outperforming other SOTA methods. In the badminton dataset, it achieved the best performance in both RMSE and accuracy, demonstrating its effectiveness across all visibility levels, from fully visible to occluded objects. These results validate TOTNet’s robust and versatile tracking capabilities across diverse scenarios.

\subsection{Ablation Studies}
To validate each component's effectiveness, we conducted an ablation study on the TTA datasets using TOTNet as the base model. Table \ref{tab:ablation_study} shows how each progressive modification improved performance.
First, the visibility-weighted BCE loss prioritized occluded and low-visibility frames, enhancing robustness under occlusion. Next, occlusion augmentation simulated diverse scenarios, forcing the model to rely on temporal and spatial context, significantly boosting accuracy in challenging conditions. Finally, integrating optical flow provided explicit motion cues, improving tracking accuracy and consistency under rapid motion or occlusion.
These components complementarily enhanced TOTNet’s performance, with the base model already outperforming the best existing methods \cite{tarashima2023widely}, highlighting its superior architecture. Additional experiments on varying input frames and using the middle frame as the target are detailed in the supplementary materials.

%% file: sec/5_conclusion.tex
\section{Conclusion}
\label{sec:conclusion}
We present TOTNet, a novel DL framework for occlusion-robust ball tracking in sports videos. Evaluated on four datasets across tennis, table tennis, and badminton, TOTNet consistently outperforms existing SOTA methods, especially under partial and full occlusions. To support real-world evaluation, we introduce the TTA dataset—an occlusion-rich, professionally annotated benchmark collected from actual Paralympic table tennis matches.

TOTNet integrates spatial and temporal cues via a 3D convolutional architecture and motion modeling modules, enabling accurate tracking even under limited visibility. While this design increases computational overhead, it is well-suited for offline applications such as post-match analysis, coaching, and referee support.

Developed in collaboration with Paralympics Australia, our work emphasizes real-world sports analytics needs. The TTA dataset reflects practical deployment scenarios, and the model offers a strong foundation for downstream tasks such as ball position tracking and event detection. Future work will focus on improving efficiency through techniques like temporal shift modules and transformer-based architectures.